\documentclass[conference]{IEEEtran}
\usepackage{cite}
\usepackage{amsmath,amssymb,amsfonts}
\usepackage{algorithmic}
\usepackage{graphicx}
\usepackage{textcomp}
\usepackage{xcolor}
\usepackage{subcaption}
\usepackage{comment}
\def\BibTeX{{\rm B\kern-.05em{\sc i\kern-.025em b}\kern-.08em
    T\kern-.1667em\lower.7ex\hbox{E}\kern-.125emX}}
\begin{document}

\title{Estimating Multiplicative Relations in Neural Networks  \\
}

\author{\IEEEauthorblockN{Bhaavan Goel}
\IEEEauthorblockA{\textit{bhaavangoel.beece14@pec.edu.in} }
}

\maketitle

\begin{abstract}
Universal approximation theorem suggests that a shallow neural network can approximate any function. The input to neurons at each layer is a weighted sum of previous layer neurons and then an activation is applied. These activation functions perform very well when the output is a linear combination of input data. When trying to learn a function which involves product of input data, the neural networks tend to overfit the data to approximate the function. In this paper we will use properties of logarithmic functions to propose a pair of activation functions which can translate products into linear expression and learn using backpropagation. We will try to generalize this approach for some complex arithmetic functions and test the accuracy on a disjoint distribution with the training set.
\end{abstract}


\section{Introduction}
The quality of a model is typically measured by its ability to generalize from a training set to previously unseen data from the similar distribution. Models based on neural network architecture with linear activation functions provide good accuracy on testing data for functions which are linear combination of inputs described by:
\[y = c+\sum\limits_i a_i x_i\]
To estimate non linear outputs, the activation functions can be carefully set to some non-linear functions. The non linear activation functions have to be differential in the relevant domain for back propagation learning\cite{b1} to work. Generally these non linear activation functions can approximate a wide range of practical problems and work very well on previously unseen real life data. However, we focus on a very specific problem of trying to approximate a product function:
\[y = c\prod\limits_i x_i\]
Multiplication cannot be generalized using some linear combination of variables, thus it becomes necessary to think of some other approach to accommodate these product functions within the existing neural networks. In this paper we will: 
\begin{enumerate}
  \item introduce customized logarithm-exponential activation pair which could learn multiplication
  \item test the accuracy on data which is disjoint with the training set
  \item compare the results with other activation functions
\end{enumerate}

The following section describes the data used in the experiments. Afterwards, we introduce our method and discuss the architecture, its training and relation to prior art. Throughout our discussion, we will refer to $\log_e$ as $\log$ for convenience.
\section{Data}

\begin{figure*}[!htb]
\minipage{0.52\textwidth}
  \includegraphics[width=\linewidth]{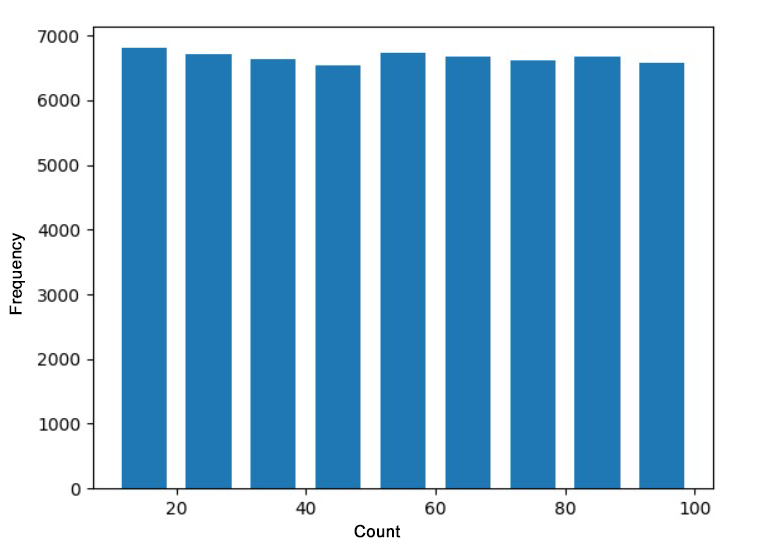}
  \caption{Training dataset}\label{fig:training}
\endminipage\hfill
\minipage{0.52\textwidth}
  \includegraphics[width=\linewidth]{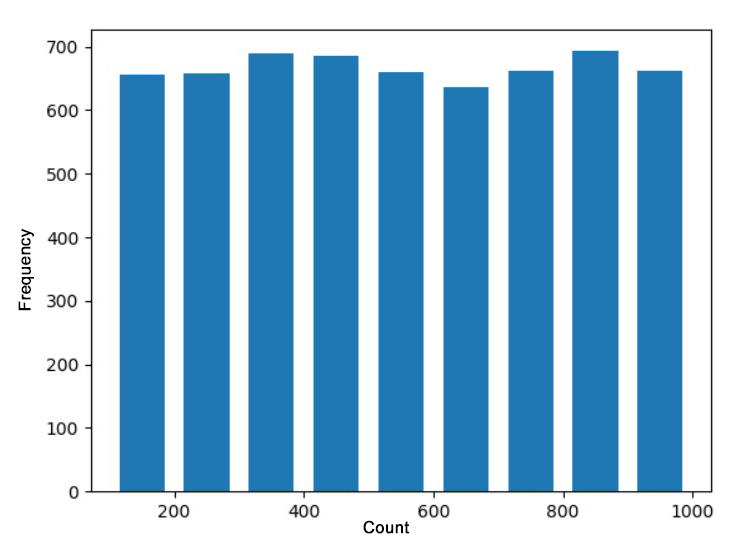}
  \caption{Testing dataset}\label{fig:testing}
\endminipage\hfill
\end{figure*}

We will generate some uniformly distributed synthetic data. We will have a set of variables $\{x_1, x_2, ..., x_n\}$ as input and a single value $y$ as output. Each input variable will be in the range $[10 ,100)$ for training and in the range $[100,1000)$ for testing. Using disjoint sets for training and testing, we will get a better picture if a model is over-fitting the training data. The histogram of the training and testing data is shown in figure \ref{fig:training} and \ref{fig:testing} respectively. As a primary objective, we will try to approximate three functions: 
 \[y = \frac{x_1 x_2}{N}\]
\[y = \frac{x_1 x_2 x_3}{N}\] 
\[y = \frac{x_1 x_2 x_3 x_4}{N}\]
and compare the results with various activation functions. Here $N = 10^n$ represents the normalizing factor to keep the scale of output $y$ comparable to the inputs $x_i$. The normalizing factor N will be kept variable to see how it impacts the accuracy of the model.
\graphicspath{ {./images/} }

\section{Symmetric Logarithm-Exponential activation pair}
The output of a product function scales very quickly compared to the individual inputs, which creates the problem of gradient explosion during backpropagation. In this paper, we try to leverage the well-known property of logarithm function:
\[\log\left(\prod\limits_i x_i\right) = \sum\limits_i \log x_i , \forall x>0\]
We will set logarithm as the activation of a layer to capture the sense of multiplication through addition of log values. Now using the property of exponential function:
\[e^{\sum\limits_i x_i} = \prod\limits_i e^{x_i} \]
we get the initial inputs as a product with their corresponding weights as exponents.
\[w_1\log x_1 + w_2 \log x_2 \xrightarrow{\exp} x_1^{w_1}*x_2^{w_2}\]
Backpropagation should update the weights $w_1$ and $w_2$ to 1 if multiplication is suitable for reducing the loss and -1 if division can reduce the loss.

\subsection{Symmetric Logarithm Unit}\label{AA}
Since $\log$ is only defined for positive inputs, we need to define how the activation function would behave for 0 and negative values. Keeping in mind the discussed property of $\log$, we will try to define the activation such that it is:
\begin{itemize}
   \item  defined for real number space $\mathbb{R}$
   \item  differentiable in the whole domain
   \item  symmetric about origin
 \end{itemize}
The first two conditions allow backpropagation to work for all the real inputs and symmetricity about origin avoids bias shift. Thus we define activation function for the first layer as:
\begin{equation}
f(x) =
\begin{cases} 
      \log(x+1) & x\geq 0 \\
      -\log(1-x) & x< 0 
   \end{cases}
\label{eq:log}
\end{equation}
The differentiability of function \ref{eq:log}  $\forall x \in \mathbb{R} -\{0\}$ follows from differentiability of $\log x$,  $\forall x>0$. 

\begin{figure}[h]
\includegraphics[width=\linewidth, height=6cm]{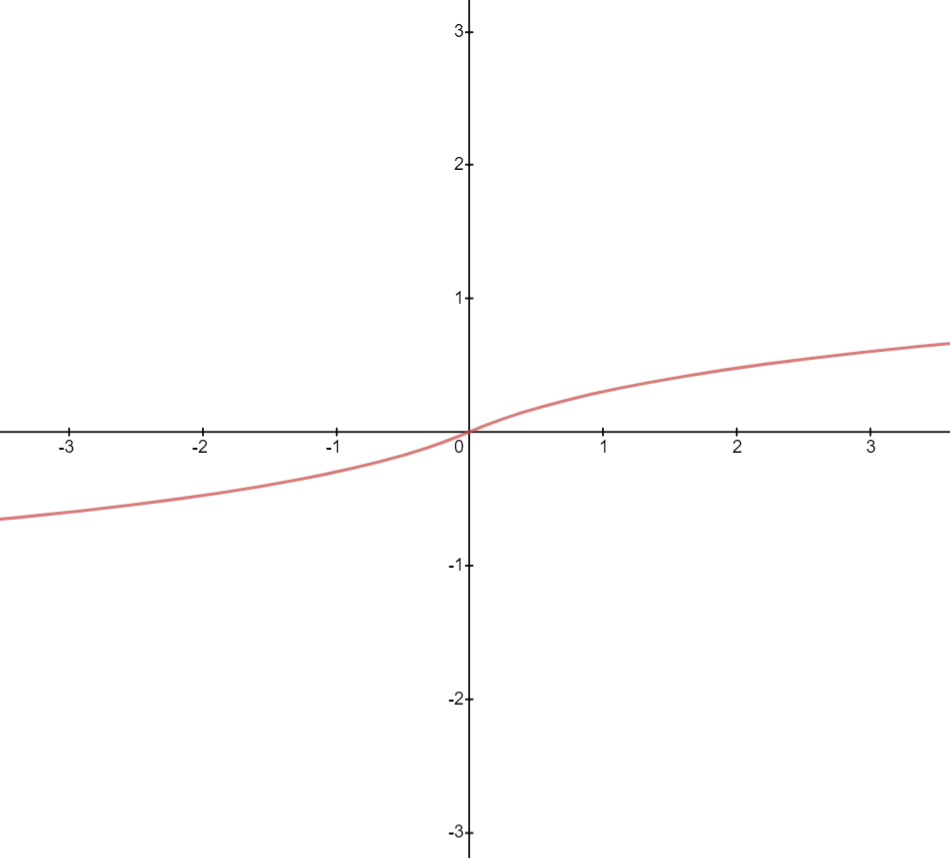}
\caption{Symmetric $\log$ function defined in eq.\ref{eq:log}}\label{fig:log}
\end{figure}

Considering the critical point $x=0$ 
\[\lim_{x=0,\delta \to 0^+} \frac{\log(x+\delta+1) - \log(x+1)}{\delta} = \frac{\log(1+\delta)}{\delta} = 1\]
\[\lim_{x=0,\delta \to 0^-} \frac{-\log(-x-\delta+1) + \log(-x+1)}{\delta} = -\frac{\log(1-\delta)}{\delta} = 1\]
shows that the function is differentiable for  $\forall x \in \mathbb{R}$.
\subsection{Symmetric Exponential Unit}\label{AA}
Similarly based on the properties discussed for Symmetric Logarithm unit, we will define the exponential activation function for the second layer which is continuous and differentiable $\forall x \in \mathbb{R}$:
\begin{equation}
f(x) =
\begin{cases} 
      e^x -1 & x\geq 0 \\
      -e^{-x} +1 & x< 0 
   \end{cases}
\label{eq:exp}
\end{equation}
\begin{figure}[h]
\includegraphics[width=\linewidth, height=6.4cm]{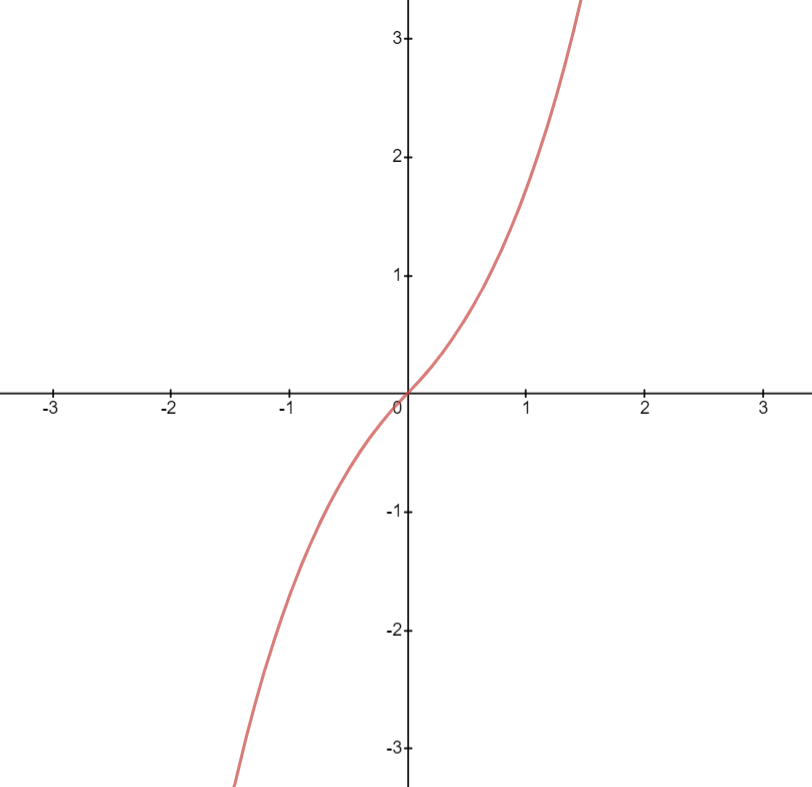}
\caption{Symmetric exponential function defined in eq.\ref{eq:exp}}\label{fig:exp}
\end{figure}

\begin{figure*}
\includegraphics[width=0.6\linewidth]{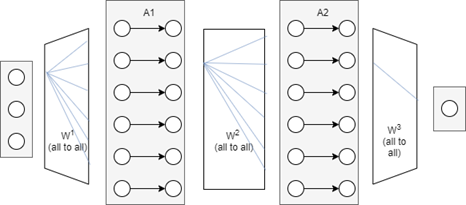}
\centering
\caption{Architecture of shallow NN model, A1 and A2 represent activation functions}\label{fig:model}
\end{figure*}

\section{Experimental Setup}
Keras\cite{b5} with Google TensorFlow backend\cite{b6} was used to implement the deep learning algorithms in this study with the aid of other scientific computing libraries: matplotlib\cite{b7}, numpy\cite{b8}, and Scikit-learn\cite{b9}. All experiments in this study were conducted on a laptop computer with Intel Core (TM) i7-9750H CPU @ 2.60GHz, 16GB of DDR3 RAM, and NVIDIA GeForce GTX 1650 4GB. 

\section{The Model}
\begin{figure}[h]
\includegraphics[width=0.75\linewidth]{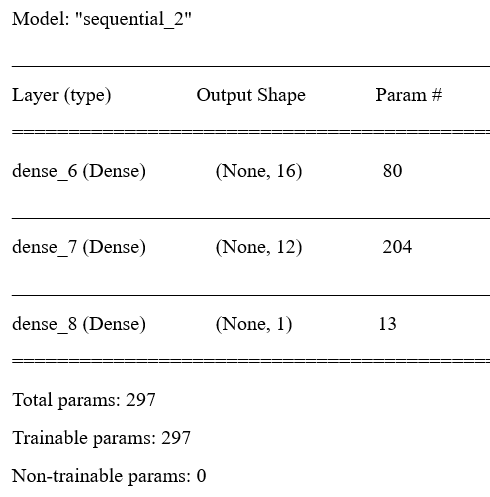}
\centering
\caption{Keras model}\label{fig:layers}
\end{figure}

We will prepare a shallow feed-forward neural network containing 2 hidden layers and test various combinations of activation functions on our prepared dataset. The final output dense layer will use linear activation which is default for Keras\cite{b5}. The abstract architecture of the model has been described in Fig. \ref{fig:model} and the implemented Keras\cite{b5} architecture in figure ~\ref{fig:layers}. For comparison, we will consider activation functions from \{elu\cite{b10}, hard sigmoid, linear, relu\cite{b12}, selu\cite{b13}, sigmoid, softmax, softplus, softsign, swish\cite{b14} and tanh\}. These 11 unique activation functions make a total of $121(11 \times 11)$ activation pairs. We will test the score of our proposed activation function against all the $121$ possible pairs.

\begin{figure*}
\begin{subfigure}{0.5\textwidth}
\includegraphics[scale=0.52]{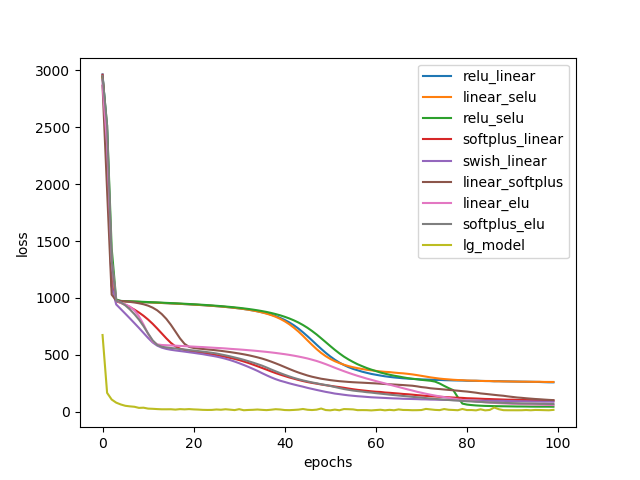}
\centering
\caption{$y=x_1 x_2$}\label{fig:train1}
\end{subfigure}
\begin{subfigure}{0.5\textwidth}
\includegraphics[scale=0.52]{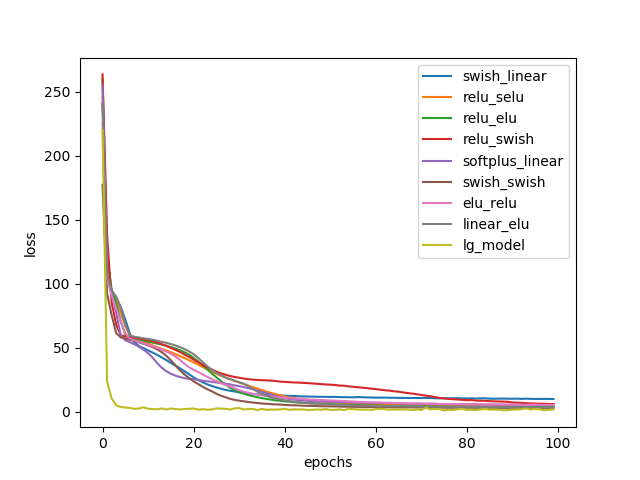}
\centering
\caption{$y=\frac{x_1 x_2}{10}$}\label{fig:train2}
\end{subfigure}
\begin{subfigure}{0.5\textwidth}
\includegraphics[scale=0.52]{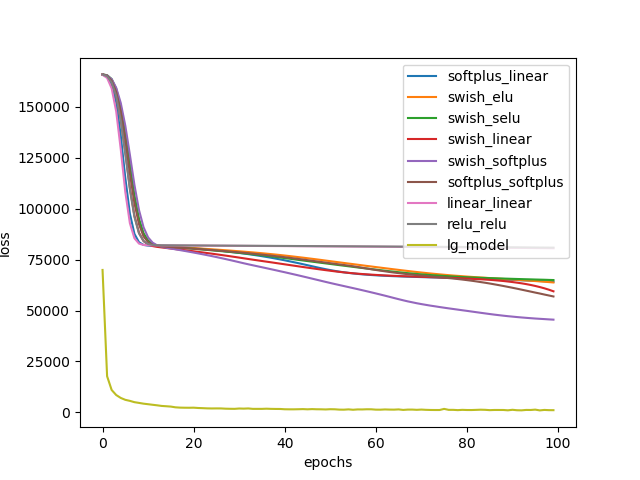}
\centering
\caption{$y=x_1 x_2 x_3$}\label{fig:train4}
\end{subfigure}
\begin{subfigure}{0.5\textwidth}
\includegraphics[scale=0.52]{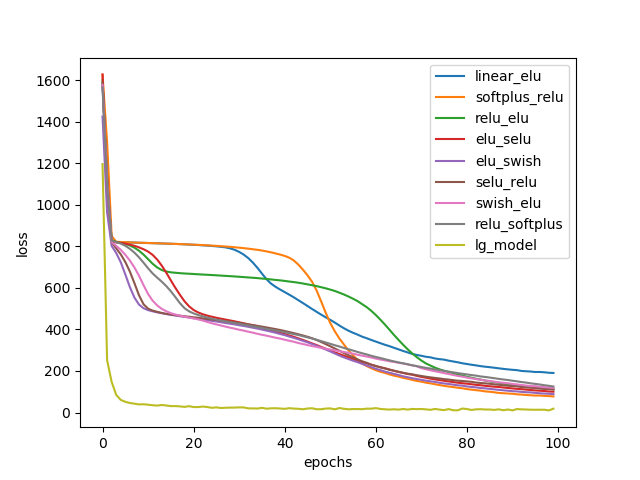}
\centering
\caption{$y=\frac{x_1 x_2 x_3}{100}$}\label{fig:train6}
\end{subfigure}
\begin{subfigure}{0.5\textwidth}
\includegraphics[scale=0.52]{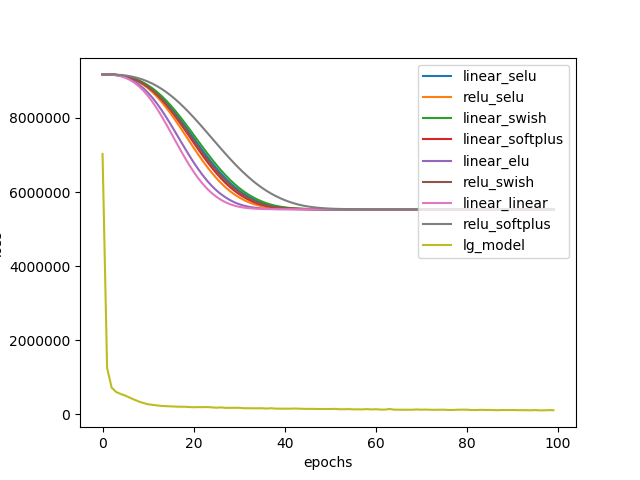}
\centering
\caption{$y=x_1 x_2 x_3 x_4$}\label{fig:train8}
\end{subfigure}
\begin{subfigure}{0.5\textwidth}
\includegraphics[scale=0.52]{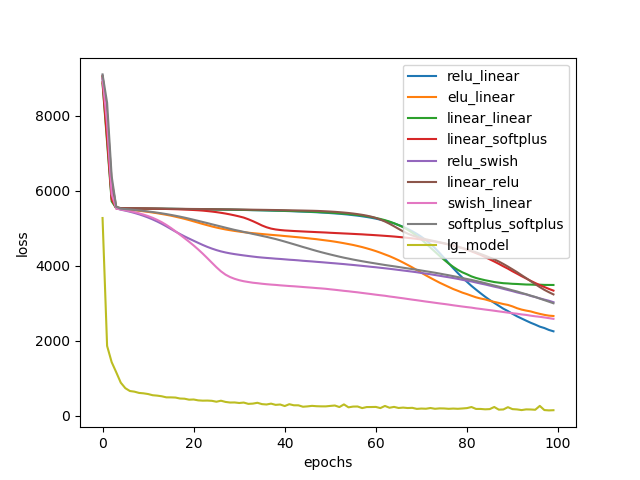}
\centering
\caption{$y=\frac{x_1 x_2 x_3 x_4}{1000}$}\label{fig:train11}
\end{subfigure}
\caption{Training loss over 100 epochs for different input size and normalization. Here "lg\_model" is our proposed model.}
\label{fig:trainplots}
\end{figure*}

\section{Results}
\subsection{Product Functions}
\subsubsection{Training Loss}
The model is trained to minimize the Mean Absolute Error (MAE) using adaptive momentum estimation (Adam)\cite{b11} optimizer for our experiments. We have trained the model for product functions with up to 4 input variables and different normalization factors. Normalization is essential to ensure that the product of inputs does not increase unboundedly. The plots in figure \ref{fig:trainplots} show $7$ (out of $121$) activation pairs with minimum training error after $100$ epochs along with our proposed logarithm – exponential activation pair. In all cases our model not only achieves the minimum training loss but also converges in minimum number of epochs. In cases where the output is not normalized to the scale of input variables, other activation functions show much larger training loss than our model.

\subsubsection{Testing Error}
We will consider the usual MAE and Mean Percentage Error ($\%_{err}$) which is calculated as:
\begin{equation}
\%_{err} = \frac{100}{n} \sum\limits_i \frac{|y^i_{pred} – y^i_{actual}|} {|y^i_{actual}|}
\end{equation}
The table depicts that the proposed model has much lower $\%_{err}$ compared to other activation pairs for the test data. It also suggests that in cases where other activation functions showed lower training error were over-fitting the training data.
\begin{table}[ht]
\caption{Testing results for different product functions}
\resizebox{\columnwidth}{1.9cm}{%
\begin{tabular}{l|l|l|l|l|}
\cline{2-5}
                                               & \multicolumn{2}{c|}{Our Model} & \multicolumn{2}{c|}{Next Best}  \\ \hline
\multicolumn{1}{|l|}{Expression}               & $ \%_{error}$    & MAE         & Activation Pair & $ \%_{error}$ \\ \hline
\multicolumn{1}{|l|}{$x_1x_2$}                 & 15.453791        & 47284.32    & relu\_linear    & 61.191017     \\ \hline
\multicolumn{1}{|l|}{$x_1x_2/10$}             & 7.801138         & 1565.0941   & relu\_selu      & 69.28548      \\ \hline
\multicolumn{1}{|l|}{$x_1x_2x_3$}           & 16.815907        & 27887974    & elu\_elu        & 98.30871      \\ \hline
\multicolumn{1}{|l|}{$x_1x_2x_3/100$}       & 24.82434         & 394495.03   & softplus\_relu  & 93.41624      \\ \hline
\multicolumn{1}{|l|}{$x_1x_2x_3x_4$}      & 28.907835        & 3.4e+10     & softplus\_selu  & 99.79         \\ \hline
\multicolumn{1}{|l|}{$x_1x_2x_3x_4/1000$} & 39.76962         & 34425464   & elu\_linear     & 99.70         \\ \hline
\end{tabular}%
}
\label{Tab:test_results}
\end{table}

\subsection{Complex function}
We will also show the effectiveness of our method for estimating more complex arithmetic functions. The function we try to estimate is:
\begin{equation}
	y = x_1(x_2 + x_3) + x_4
\label{eqn:complex}
\end{equation}
\subsubsection{Training Loss}
\begin{center}
\begin{figure}[h]
\includegraphics[width=0.8\linewidth]{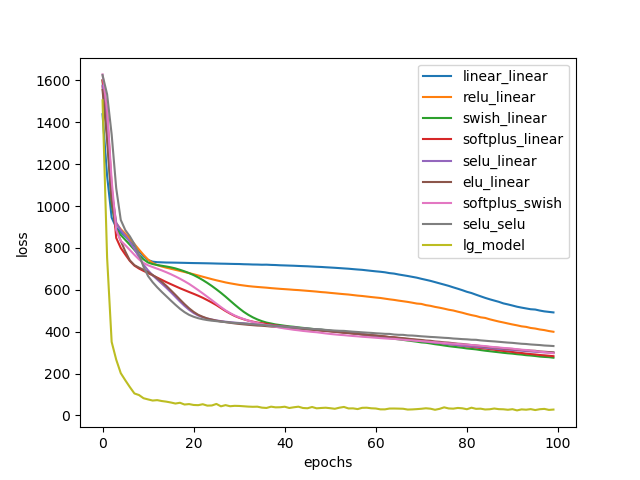}
\centering
\caption{Training loss for $y = x_1(x_2 + x_3) + x_4$. Here "lg\_model" is our proposed model}
\label{fig:complex_loss}
\end{figure}
\end{center}

The training loss for the function described by eq.\ref{eqn:complex} is also minimum for our model and also converges quickly. The training loss is depicted in figure \ref{fig:complex_loss}.
\subsubsection{Testing Error}
The mean percentage error without any architectural changes for the complex formula is 22.794909. which is much less than the next best 97.09597 for selu-linear activation pair.
\section{Related Work}
Durbin and Rumelhart, 1989 \cite{b3} tries to achieve the results of product function by taking weighted product of inputs. This approach achieves desired results but requires changes in the commonly used training loop where we generally take weighted sum of the inputs. Georg Martius and Christoph H. Lampert \cite{b4} propose a cell that takes two inputs at a time and calculates their product. It also performs a weighted product of the inputs. NLRELU \cite{b2} introduces an activation similar to the activation function used in the first layer of our model. NLRELU has various advantages but it alone cannot estimate a product function very well.

\section{Conclusion}
We presented and studied the behavior of logarithm paired with an exponential activation function for estimating product functions. For data which have multiplicative relations between the inputs, we saw that our customized activation pair achieved minimum scoring error compared to other activation functions and also avoids over-fitting on the training data with a very shallow feed-forward neural network. We proved the differentiability of our functions which allows end-to-end training using backpropagation and thus can be integrated within deep neural networks to estimate hidden product relations.

\section{Acknowledgement}
We thank Soham Pal and Shivam Gupta for continuous encouragement and feedback throughout this paper.

\vspace{12pt}

\end{document}